\documentclass[letterpaper, 10 pt, journal, twoside]{IEEEtran}

\usepackage[noend]{algpseudocode}
\usepackage{algorithm}

\newcommand{\ignore}[1]{}

\newcommand{\norm}[1]{\left\Vert#1\right\Vert} 
\newcommand{\mc}[1]{\mathcal{#1}}

\newcommand{\bma}[1]{\left[\begin{array}{ #1}}
\newcommand{\ema}{\end{array}\right]}

\DeclareMathAlphabet{\mbf}{OT1}{ptm}{b}{n}
\newcommand{\mbs}[1]{{\boldsymbol{#1}}}

\newcommand{\mbsdot}[1]{{\dot{\boldsymbol{#1}}}}

\newcommand{\mbshat}[1]{{\hat{\boldsymbol{#1}}}}

\newcommand{\mbstilde}[1]{{\tilde{\boldsymbol{#1}}}}
\newcommand{\mbfdot}[1]{{\dot{\mbf{#1}}}}

\newcommand{\mbfhat}[1]{{\hat{\mbf{#1}}}}
\newcommand{\mbfcheck}[1]{\ensuremath{\check{\mbf{#1}}}}

\newcommand{\mbfdel}[1]{{\delta{\mbf{#1}}}}
\newcommand{\mbftilde}[1]{{\tilde{\mbf{#1}}}}

\def\fdotb{{\raisebox{-0.6ex}{ \kern0.2ex\raisebox{0.8ex}{\tiny $\hspace*{-1ex}\circ$}}}}
\def\fddotb{{\raisebox{-0.6ex}{ \kern0.2ex\raisebox{0.8ex}{\tiny $\hspace*{-1ex}\circ\circ$}}}}

\newcommand{\f}{\frac}

\newcommand{\trans}{{\ensuremath{\mathsf{T}}}} 
\newcommand{\utimes}{ {\raisebox{-0.6ex}{ \kern-1.0ex\raisebox{0.6ex}{ \small $\mathsf{v}$}}} } %
\newcommand{\beq}{\begin{equation}}
\newcommand{\eeq}{\end{equation}}
\newcommand{\bdis}{\begin{displaymath}}
\newcommand{\edis}{\end{displaymath}}
\newcommand{\beqarray}{\begin{eqnarray}}
\newcommand{\eeqarray}{\end{eqnarray}}
\newcommand{\beqarraynn}{\begin{eqnarray*}}
\newcommand{\eeqarraynn}{\end{eqnarray*}}
\newcommand{\balign}{\begin{align}}
\newcommand{\ealign}{\end{align}}
\newcommand{\balignnn}{\begin{align*}}
\newcommand{\ealignnn}{\end{align}}

\renewcommand{\theenumii}{\arabic{enumii}}
\makeatletter
\renewcommand{\p@enumii}{\theenumi.}
\makeatother

\usepackage{arydshln}
\usepackage{mathtools}

\usepackage{amsmath} 
\usepackage{amssymb}
\usepackage{float}

\usepackage{multirow}

\usepackage{capt-of}

\usepackage{cite}
\begin{document}

\title{Ultra-Wideband Teach and Repeat}

\author{Mohammed Ayman Shalaby, Charles Champagne Cossette, Jerome Le Ny, James Richard Forbes%
\thanks{This work was supported by FRQNT under grant 2018-PR-253646, the William Dawson Scholar program, the NSERC Discovery Grant program, and the CFI JELF program.} 
\thanks{M. A. Shalaby, C. C. Cossette, and J. R. Forbes are with the department of Mechanical Engineering, McGill University, Montreal, QC H3A 0C3, Canada. {\tt\footnotesize mohammed.shalaby@mail.mcgill.ca, charles.cossette@mail.mcgill.ca, james.richard.forbes@mcgill.ca.}}%
\thanks{J. Le Ny is with the department of Electrical Engineering, Polytechnique Montreal, Montreal, QC H3T 1J4, Canada. {\tt\footnotesize jerome.le-ny@polymtl.ca.}}
}


\maketitle

\begin{abstract}
Autonomously retracing a manually-taught path is desirable for many applications, and Teach and Repeat (T\&R) algorithms present an approach that is suitable for long-range autonomy. In this paper, ultra-wideband (UWB) ranging-based T\&R is proposed for vehicles with limited resources. By fixing single UWB transceivers at far-apart unknown locations in an indoor environment, a robot with 3 UWB transceivers builds a locally consistent map during the teach pass by fusing the range measurements under a custom ranging protocol with an on-board IMU and height measurements. The robot then uses information from the teach pass to retrace the same trajectory autonomously. The proposed ranging protocol and T\&R algorithm are validated in simulation, where it is shown that the robot can successfully retrace the taught trajectory with sub-metre tracking error. 
\end{abstract}


\IEEEpeerreviewmaketitle

\section{Introduction} \label{sec:intro}

Autonomy for navigation in GPS-denied environments has seen significant advancements in recent years due to increased focus from the robotics community. The research efforts are backed by increased emphasis in developing reliable and real-time solutions to address real-world problems, such as warehouse operations or indoor inspections. In order to facilitate scalability of navigation systems for long-range autonomy and multi-robot applications, computational and financial costs must be reduced. This motivated the recent surge in using UWB signals as an alternative to demanding exteroceptive sensors such as stereo cameras and LIDAR scanners. 

UWB signals are used in robotic applications to measure the distance or \emph{range} between two UWB transceivers \cite{sahinoglu2008}. The range measurement is most commonly computed from the time-of-flight (ToF) of signals transmitted from one transceiver and received at another. UWB-based positioning typically involves having a few static transceivers, or \emph{anchors}, at fixed known locations communicating with one or more mobile transceivers, or \emph{tags}, in order to obtain range measurements between them. This then allows positioning of the robot through trilateration or filtering \cite{Cano2019, Mueller2015a, Ledergerber2015}. 

\begin{figure}
    \centering
    \includegraphics[trim=13cm 3cm 13cm 3cm, clip, width=\columnwidth]{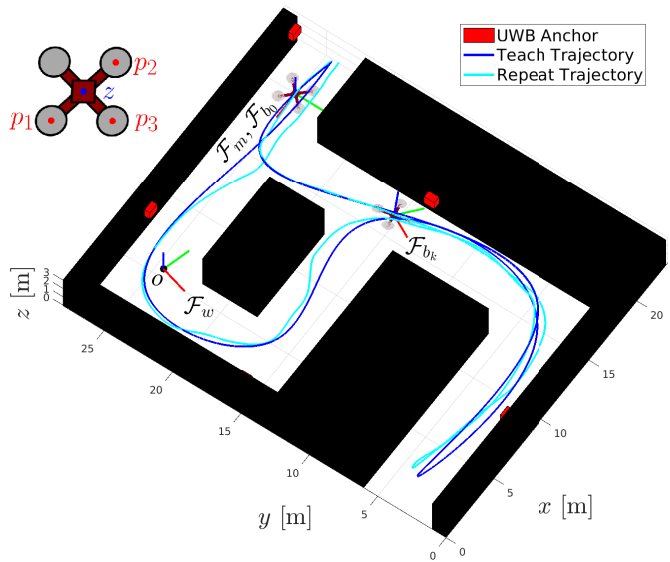}
    \caption{A 3D indoor environment with spaced-out UWB anchors fixed at unknown positions. The robot is equipped with 3 UWB ranging tags, and successfully retraces the teach trajectory autonomously in the repeat pass.}
    \label{fig:env}
\end{figure}

Alternatively, to overcome the need for a fixed infrastructure, the most common approach for indoor navigation is simultaneous localization and mapping (SLAM). Most SLAM algorithms are vision- \cite{Engel2013, Engel2015} or LIDAR-based \cite{Droeschel2018}, which requires computationally-heavy processing and large storage to process and store images or point clouds. Additionally, SLAM is oftentimes restricted to small areas, as global map consistency is only achieved through complex data association and loop closure algorithms that are difficult for larger maps and not always possible \cite{Campos2021}. Range-based SLAM has also been proposed mainly for fixed infrastructure \cite{Cao2020a, Cao2020b}, while \cite{Funabiki2021} proposes dropping beacons from a quadcopter and utilizing the associated range measurements in a SLAM framework. 

A more recent approach to overcome the pitfalls of SLAM is T\&R, where a robot learns a locally consistent map of the environment during a teach pass, and then retraces the trajectory it took during a repeat pass using the learnt map \cite{Furgale2010}. As long as a robot behaves the same way every time relative to its local features, the robot is able to retrace the same trajectory \cite{Furgale2010, Krajnik2018}. This has been studied using stereo cameras \cite{Furgale2010} and LIDAR \cite{McManus2011} on ground vehicles, and also with monocular cameras \cite{Warren2018} and aerial vehicles \cite{Nitsche2019}. As with all vision-based algorithms, difficulties arise when the environment changes due to weather or other external factors. This requires relying on colour-constant imaging \cite{Paton2015}, using multiple experiences simultaneously \cite{Paton2016}, and/or deep learning approaches for image association and localization \cite{Camara2020, Gridseth2020}.

The contributions of this paper are threefold and revolve around combining the advantages of using a T\&R framework with those of using UWB-based range measurements for localization. Firstly, a ranging protocol is proposed to allow a 3-tag agent to synchronize the clocks of its tags and receive 3 range measurements from a fixed anchor with only one two-way ranging (TWR) transaction. This therefore allows necessary additional information for localization \cite{Shalaby2021} without compromising the frequency in which range measurements are recorded. Secondly, a novel UWB-based T\&R framework is presented for a vehicle moving in an environment with spaced-out fixed anchors at unknown locations over a large area. The robot also utilizes an inertial measurement unit (IMU) and height measurements. As is standard with T\&R, the robot is manually controlled over a trajectory that is then autonomously retraced. Lastly, the presented framework is tested in simulation using the environment shown in Figure \ref{fig:env} and is shown to achieve sub-metre tracking performance.  

The rest of this paper is structured as follows. The notation is summarized and the problem is formulated in Section \ref{sec:prob_form}. The proposed teach and repeat passes are covered in Sections \ref{sec:teach_pass} and \ref{sec:repeat_pass}, respectively. The results are then discussed in Section \ref{sec:results}, and some concluding remarks are given in Section \ref{sec:conclusion}. 




    



 

\section{Problem Formulation} \label{sec:prob_form}

\subsection{Notation}


Anchor $i$ is located at point $a_i$, and the set of all static anchors is denoted $\mc{A}$. Tag $i$ on the robot is located at point $p_i$, and the IMU at point $z$, as shown in Figure \ref{fig:env}. The subscript $k$ is used throughout the text to denote the $k^\text{th}$ time-step, while a subscript $i:i+j$ denotes all the variables at time-steps $k \in \{ i, \ldots, i+j \}$. Meanwhile, $K$ is used to denote the total number of time-steps in each of the teach and repeat passes. When applicable, the subscript $0:K$ is omitted for conciseness when addressing the full trajectory. The superscripts $\mathfrak{t}$, $\mathfrak{r}$, and $\mathfrak{i}$ are used to allocate variables to the teach pass, the repeat pass, and the initialization phase of the repeat pass, respectively.

The reference frames used in this text are shown in Figure \ref{fig:env}. A local reference frame, the \emph{world} frame, is denoted $\mc{F}_w$, while the body-fixed reference frame is denoted $\mc{F}_{b_k}$ at time-step $k$. The map frame $\mc{F}_m$ is fixed to the body frame at $k=0$ (i.e., $\mc{F}_m = \mc{F}_{b_0}$), and differs from $\mc{F}_w$ only through a rotation about the $z$-axis and a translation about the $x$- and $y$-axes.

The position of a point $i$ relative to another point $j$ resolved in the frame $\mc{F}_q$ in $\mathbb{R}^3$ is denoted with $\mbf{r}_q^{ij}$. The corresponding velocity with respect to frame $\mc{F}_\ell$ is denoted with $\mbf{v}_q^{ij / \ell}$. The rotation vector $\mbs{\phi}_{\ell q}$ is associated with the direction cosine matrix (DCM) $\mbf{C}_{\ell q} \in SO(3)$, representing the rotation from $\mc{F}_q$ to $\mc{F}_\ell$. Meanwhile, $\mbftilde{r}_q^{ij}$ is used to represent a position in $\mathbb{R}^2$, and the heading $\tilde{\theta}_{\ell q}$ is associated with the DCM $\mbftilde{C}_{\ell q} \in SO(2)$.

The estimate or posterior mean of a random variable is denoted with $\hat{( \cdot )}$, while a prior or prediction without measurement correction is denoted with $\check{ ( \cdot )}$. Meanwhile, $(\cdot)^\times$ denotes the skew-symmetric cross-product matrix operator in $\mathbb{R}^3$, and $\norm{\cdot}_\mbs{\Sigma} = \sqrt{(\cdot)^\trans \mbs{\Sigma}^{-1} (\cdot)}$ denotes the weighted Euclidean norm. Lastly, $\mbf{w} \sim \mc{N} \left( \mbf{0}, \mbf{Q} \right)$ and $\mbs{\nu} \sim \mc{N} \left( \mbf{0}, \mbf{R} \right)$ represent process and measurement Gaussian white noise, respectively. The random variables and their covariance matrices always share the same subscripts and superscripts \big(e.g., $\mbf{w}_{b_k}^{\text{acc},\mathfrak{t}} \sim \mc{N} (0, \mbf{Q}_{b_k}^{\text{acc},\mathfrak{t}})$\big). 

\subsection{Overview}

\begin{figure}
    \centering
    \includegraphics[width=0.8\columnwidth]{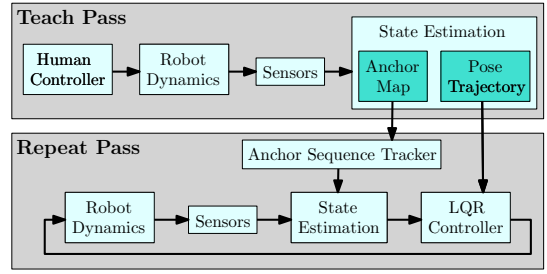}
    \caption{A high-level overview of the major processing blocks of UWB T\&R.}
    \label{fig:block_diagram}
\end{figure}

Consider a robot in an environment with $n$ spaced-out and fixed UWB anchors at unknown locations $\mbf{r}_w^{a_i o}$, $i \in \{1, \ldots, n\}$, as shown in Figure \ref{fig:env}. Initially, the robot is manually driven on a trajectory denoted $\mbs{\pi}^\mathfrak{t}$ that has approximately the same starting and ending point on a flat ground. As the robot moves, it records measurements from an IMU, a height sensor, and range measurements with any anchor that lies within communication range. The robot is tasked with estimating the trajectory it is following such that it can be autonomously repeated in the future. Additionally, the robot maps the anchors as they are detected by estimating their position. These two tasks consequently fall under what is termed the teach pass, as the robot learns during this process the trajectory it must follow and a map of the environment it encounters, which is just a map of the UWB anchor positions. 

Despite having access to only the estimated teach trajectory $\mbshat{\pi}^\mathfrak{t}$ and the estimated anchor positions, the robot then attempts to repeat the teach trajectory $\mbs{\pi}^\mathfrak{t}$ autonomously. This is termed the repeat pass. The robot first estimates its pose relative to the initial pose in the teach pass, since the robot at the end of the teach trajectory does not necessarily land at the same exact position and with the same heading. The repeat pass then involves using a state estimator and a trajectory-tracking controller to retrace the teach trajectory. A summary of the major processing blocks is given in Figure \ref{fig:block_diagram}.


As is common with SLAM approaches, the robot estimates the trajectory relative to its initial pose. Therefore, at any time-step $k \in \{1, \ldots, K\}$ in the teach or repeat pass, the trajectory states to be estimated are
\begin{equation}
    \mbs{\pi}_k = \big[ \begin{array}{ccc}
    (\mbf{r}_m^{z_ko})^\trans & (\mbf{v}_m^{z_ko/w})^\trans & (\mbs{\phi}_{b_k m})^\trans
    \end{array} \big]^\trans. 
\end{equation}
In the teach pass, the estimated trajectory is stored alongside the estimated anchor positions, while in the repeat pass the estimated state is input to the controller. 

The continuous-time kinematics of the trajectory are
\begin{align}
    \mbfdot{r}_m^{zo} &= \mbf{v}_m^{zo/w}, \label{eq:process_model_1} \\
    \mbfdot{v}_m^{zo/w} &= \mbf{C}_{mb} \left( \mbf{u}^{\text{acc}}_b + \mbs{\beta}_b^\text{acc} + \mbf{w}_b^\text{acc} \right) + \mbf{g}_m, \\
    \mbfdot{C}_{mb} &= \mbf{C}_{mb} \left( \mbf{u}^{\text{gyr}}_b + \mbs{\beta}_b^\text{gyr} + \mbf{w}_b^\text{gyr} \right)^\times,
\end{align}
where $\mbf{u}_b^\text{acc}$, $\mbs{\beta}_b^\text{acc}$, and $\mbf{w}_b^\text{acc}$ denote the accelerometer measurement, bias, and Gaussian white noise, respectively, and $\mbf{u}_b^\text{gyr}$, $\mbs{\beta}_b^\text{gyr}$, and $\mbf{w}_b^\text{gyr}$ denote the gyroscope measurement, bias, and Gaussian white noise, respectively. Meanwhile, $\mbf{g}_m$ is the gravity vector resolved in the map frame. Assuming that the robot lies on a flat ground at the beginning of the teach pass and that $\mbf{g}_w = \left[ \begin{array}{ccc}
    0 & 0 & -g
\end{array} \right]$, then $\mbf{g}_m = \mbf{g}_w$.

In addition to the trajectory states, bias states must be estimated to correct the IMU measurements, and clock states must be estimated to correct the UWB range measurements. The accelerometer and gyroscope suffer from slowly-changing biases that can significantly degrade the estimation performance if not properly addressed. The evolution of the biases is modelled as a random walk, 
\begin{align}
    \mbsdot{\beta}_b^\text{acc} = \mbf{w}_b^{\beta,\text{acc}}, \qquad
    \mbsdot{\beta}_b^\text{gyr} = \mbf{w}_b^{\beta,\text{gyr}}. \label{eq:process_model_biases}
\end{align}

Meanwhile, the communicating UWB tags rely on ToF measurements to calculate distance, but suffer from having unsynchronized clocks. To overcome this issue, this work proposes a TWR procedure between Tag 1 on the agent and the relevant anchor. Meanwhile, Tags 2 and 3 ``eavesdrop" on the transmitted TWR messages, as further discussed in Section~\ref{sec:ranging}. Therefore, Tags 2 and 3 must estimate their clock offset and \emph{skew}, the time rate of change of the clock offset, relative to Tag 1, denoted by $\tau^{p_ip_1}_k$ and $\gamma^{p_ip_1}_k$, respectively for $i \in \{2, 3\}$. The process model of the clock states is modelled as 
\begin{align}
    \dot{\tau}^{p_i p_1} = \gamma^{p_i p_1} + w^{\tau, p_i p_1}, \qquad \dot{\gamma}^{p_i p_1} = w^{\gamma, p_i p_1}. \label{eq:process_model_clocks}
\end{align}
Both the biases and clock states are estimated during the teach pass and the repeat pass, but neither are recorded as part of the trajectory or fed into the controller. The sole purpose of estimating these states is to correct the IMU and range measurements to mitigate their effect on the performance of the estimators. The state being estimated at time-step $k$ is then
\begin{equation*}
    \mbf{x}_k = \big[ \begin{array}{ccccccc}
        \mbs{\pi}_k^\trans & (\mbs{\beta}_k^\text{acc})^\trans & (\mbs{\beta}_k^\text{gyr})^\trans & (\mbf{x}_k^\text{UWB})^\trans
    \end{array} \big]^\trans.
\end{equation*}
where $\mbf{x}_k^\text{UWB} = [\tau_k^{p_2p_1} \hspace{4pt} \tau_k^{p_3p_1} \hspace{4pt} \gamma_k^{p_2p_1} \hspace{4pt} \gamma_k^{p_3p_1}]^\trans$. In order to estimate $\mbf{x}_k$, two exteroceptive sensors are available. Firstly, height measurements are modelled as 
\begin{equation}
    \label{eq:height_meas}
    y^\text{height}_k = [ \begin{array}{ccc}
        0 & 0 & 1 
    \end{array} ] \mbf{r}_m^{z_k o} + \nu_k^{\text{height}}.
\end{equation}
Range measurements and the ranging protocol associated with the UWB tags are discussed in Section \ref{sec:ranging}.

\section{Ranging Protocol} \label{sec:ranging}

\begin{figure}
    \centering
    \includegraphics[width=0.8\columnwidth]{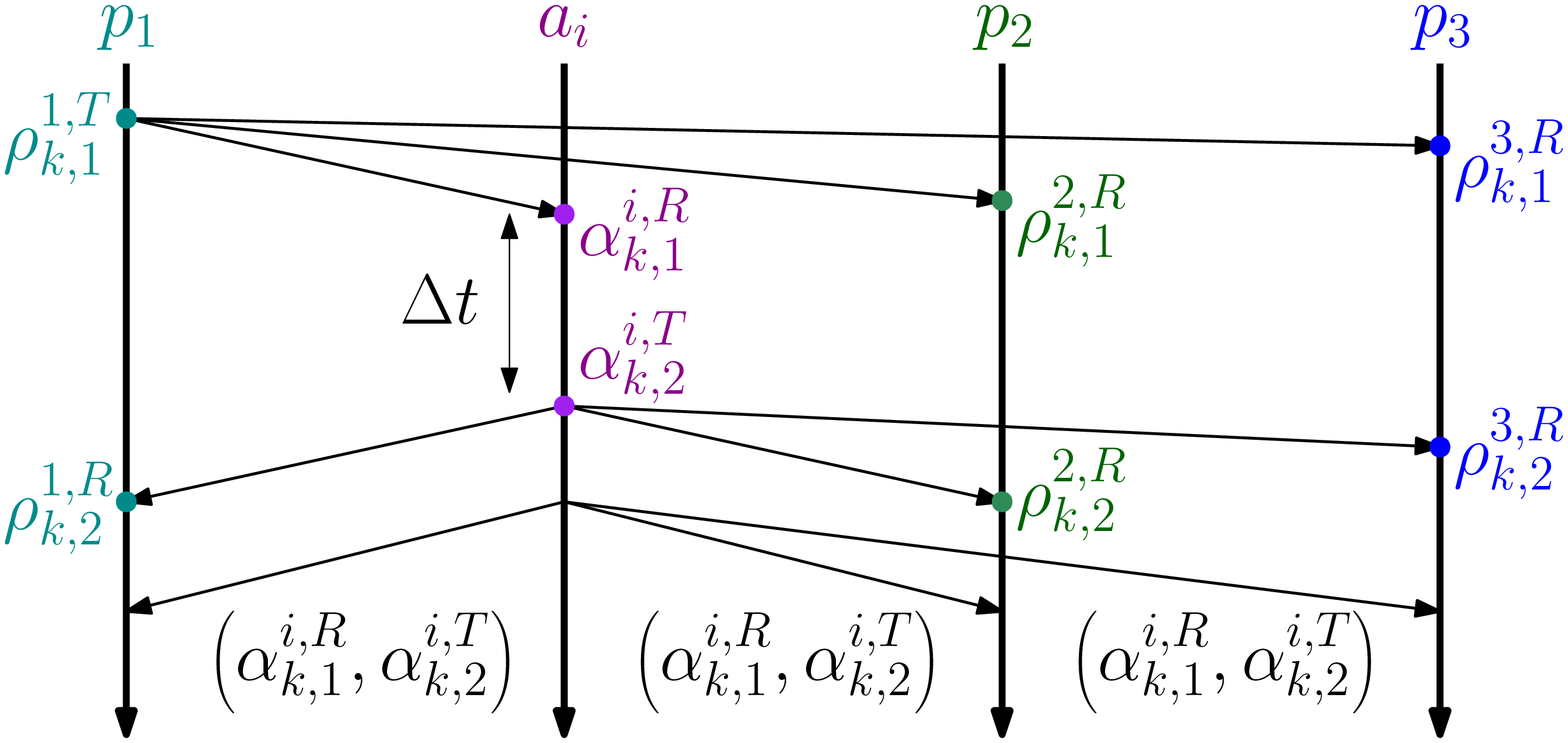}
    \caption{A schematic of the timestamps recorded by the three tags on the robot while Tag $1$ performs TWR with Anchor $i$. The notation $\rho^j_{k,\ell}$ and $\alpha^i_{k,\ell}$ are used for the $\ell^\text{th}$ timestamp recorded by the $j^\text{th}$ tag and $i^\text{th}$ anchor, respectively, at time-step $k$. Superscripts $T$ and $R$ are used to represent a transmitted and received signal, respectively. No timestamps are recorded for the last signal, it is just used to transmit information as is common in standard TWR.}
    \label{fig:ranging_timeline}
\end{figure}

In order to ensure observability of the anchor positions relative to the robot, three tags are placed on the robot \cite{sahinoglu2008, Shalaby2021}. When Anchor $i$ is within communication range with the robot, the proposed ranging protocol is shown in Figure \ref{fig:ranging_timeline}. This involves Tag $1$ performing TWR with Anchor $i$, while Tags $2$ and $3$ listen in on all signals. Therefore, the 8 recorded timestamps can be modelled as a function of the distances and clock offsets between the tags and the anchors. The timestamps for Tag $1$ and Anchor $i$ are
\begin{align}
    \rho_{k,1}^{1,T} &\triangleq t_k, \qquad \alpha_{k,2}^{i,T} = \alpha_{k,1}^{i,R} + \Delta t, \label{eq:first_timestamp} \\
    \alpha_{k,1}^{i,R} &= \rho_{k,1}^{1,T} + \f{1}{c} \norm{ \mbf{r}_m^{z_ko} - \mbf{r}_m^{a_i o} + \mbf{C}_{mb_k} \mbf{r}_{b_k}^{p_1z} } - \tau^{p_1a_i}_k + \nu_{k,1}^{1,R}, \\
    \rho_{k,2}^{1,R} &= \alpha_{k,2}^{i,T} + \f{1}{c} \norm{ \mbf{r}_m^{z_ko} - \mbf{r}_m^{a_i o} + \mbf{C}_{mb_k} \mbf{r}_{b_k}^{p_1z} } + \tau^{p_1a_i}_k + \nu_{k,2}^{1,R},
\end{align}
where $t_k$ is the timestamp in the robot's clock at the beginning of the ranging at time-step $k$, $\mbf{r}_{b_k}^{p_1z}$ is known, and $c$ is the speed of light. Meanwhile, timestamps for Tag $j \in \{2, 3\}$ are
\begin{align}
    \rho_{k,1}^{j,R} &= \rho_{k,1}^{1,T} + \f{1}{c} \norm{\mbf{r}_{b_k}^{p_jz} - \mbf{r}_{b_k}^{p_1z}} + \tau^{p_jp_1}_k + \nu_{k,1}^{j,R}, \\
    \rho_{k,2}^{j,R} &= \alpha_{k,2}^{i,T} + \f{1}{c} \norm{ \mbf{r}_m^{z_ko} - \mbf{r}_m^{a_i o} + \mbf{C}_{mb_k} \mbf{r}_{b_k}^{p_jz} } \nonumber \\ &\hspace{100pt} + \tau^{p_jp_1}_k + \tau^{p_1a_i}_k + \nu_{k,2}^{j,R}. \label{eq:last_timestamp}
\end{align}
Using relevant timestamps, a ToF measurement computed as
\begin{equation}
    \epsilon_k^{p_1a_i} = \f{1}{2}\big(\big( \alpha_{k,1}^{i,R} - \rho_{k,1}^{1,T} \big) + \big( \rho_{k,2}^{1,R} - \alpha_{k,2}^{i,T} \big)\big) \label{eq:first_tof_measurement}
\end{equation}
is standard in TWR and is of this form so that the clock offset between the anchor and the tag cancels out. In addition, the following ToF measurements are proposed for Tag $j \in \{2,3\}$,
\begin{align}
    \epsilon_k^{p_jp_1} &= \rho_{k,1}^{j,R} - \rho_{k,1}^{1,T}, \\
    \epsilon_k^{p_ja_i} &= \rho_{k,2}^{j,R} - \alpha_{k,2}^{i,T} + \alpha_{k,1}^{i,R} - \rho_{k,1}^{1,T} - \epsilon_k^{p_1a_i}, \label{eq:last_tof_measurement}
\end{align}
where clock offsets with the anchors have also been cancelled. Lastly, the range measurement model can then be derived by substituting Eqs. \eqref{eq:first_timestamp}-\eqref{eq:last_timestamp} into Eqs. \eqref{eq:first_tof_measurement}-\eqref{eq:last_tof_measurement}. This gives 2 anchor-independent ToF measurements
\begin{align}
    y_k^{p_jp_1} &= \f{1}{c} \norm{\mbf{r}_b^{p_jz} - \mbf{r}_b^{p_1z}} + \tau_k^{p_jp_1} + \nu_{k,1}^{j,R}, \label{eq:range_meas_j1}
\end{align}
and 3 anchor-dependent ToF measurements
\begin{align}
    y_k^{p_1a_i} &= \f{1}{c} \norm{ \mbf{r}_m^{z_ko} - \mbf{r}_m^{a_io} + \mbf{C}_{mb_k} \mbf{r}_b^{p_1z}} + \f{1}{2} \nu_{k,1}^{1,R} + \f{1}{2} \nu_{k,2}^{1,R}, \label{eq:range_meas_1s} \\
    y_k^{p_ja_i} &= \f{1}{c} \norm{ \mbf{r}_m^{z_ko} - \mbf{r}_m^{a_io} + \mbf{C}_{mb_k} \mbf{r}_b^{p_jz}} \nonumber \\ &\hspace{30pt}+ \tau_k^{p_j p_1} + \nu_{k,2}^{j,R} + \f{1}{2}\nu_{k,1}^{1,R} - \f{1}{2}\nu_{k,2}^{1,R}, \label{eq:range_meas_js}
\end{align}
for $j \in \{2,3\}$. Note that clock offsets with the anchors are cancelled out, and therefore none of the ToF measurements are dependent on these clock offsets. In fact, the only clock offsets that must be considered are directly measured by \eqref{eq:range_meas_j1}, since the distance between tags on the robot is known. It is also important to point out that the modelled range measurements share noise random variables, and are thus correlated.

\section{Teach Pass} \label{sec:teach_pass}


Given that the position of the anchors is unknown, the anchors must be localized when encountered in the teach pass. Characteristic of T\&R algorithms is the lack of loop closure enforcements. When the robot encounters an anchor $i \in \mc{A}$ for the first time, it estimates and stores in memory the anchor position. However, when the robot encounters the same anchor again in the future, the previously estimated position is not used, as this constitutes a loop closure that requires correcting all the states estimated between the two instances where the same anchor is detected. In a T\&R scenario, the robot disregards the fact that this anchor was previously encountered and re-localizes the anchor position, since global consistency of the map is not necessary. 

The robot tracks the sequence of initialized anchors and each anchor's ID and estimated position. Letting the $\ell^\text{th}$ encountered anchor be Anchor $i$, the initialization result is encoded in an ordered triple $\big( \mbfhat{r}_m^{\ell o}, i, \ell \big)$. The estimated anchor map is then represented as a set of ordered triples
\begin{equation*}
    \mc{M} = \{\big( \mbfhat{r}_m^{\ell o}, i, \ell \big) \vert \mbfhat{r}_m^{\ell o} \in \mathbb{R}^3, i \in \mc{A}, \ell \in \mathbb{N}^+ \}.
\end{equation*}

During the teach pass, the robot keeps track of the set of \textit{currently active anchors} $\mc{C} \subset \mc{A}$, the anchors that lie within communication range of the robot. This allows the robot to use recently-localized anchors to correct its state estimate. When an active Anchor $i$ is no longer within communication range, Anchor $i$ is dropped from the set of active anchors such that when Anchor $i$ is detected again, it is re-initialized. The teach pass sequence tracker is summarized in Algorithm 1.

\begin{algorithm} \label{alg:teach_sequence_tracker}
\caption{Teach pass anchor sequence tracker. $id$ is the ID of the communicating anchor, $\mc{C}$ is the current set of active anchors, and $\mc{M}$ is the anchor map. For all elements $(\mbf{r}, i, \ell) \in \mc{M}$, the term ``most recent" corresponds to the element with the highest $\ell$ among the elements with $i = id$.}
\begin{algorithmic}[1]
\Function{teachSequenceTracker}{$anchor$, $\mc{C}$, $\mc{M}$}

\If {$anchor \in \mc{C}$}
    
    \State $\mbfhat{r}_m^{\ell o}$ = \Call{getMostRecentPosition}{$anchor$, $\mc{M}$}
    
\Else

    \State $\mbfhat{r}_m^{\ell o}$ = \Call{initializeAnchor}{$anchor$}

    \State $n_\mc{M}$ = number of elements in $\mc{M}$

    \State $\mc{M} = \mc{M} \cup \{\mbfhat{r}_m^{\ell o}, anchor, n_\mc{M}+1\}$
    
    \State $\mc{C} = \mc{C} \cup \{anchor\}$
    
\EndIf

\For{$j \in \mc{C}$}

    \If{$j$ no longer within communication range}
    
        \State $\mc{C} = \mc{C} \backslash \{j\}$
    
    \EndIf

\EndFor

\State \Return $\mbfhat{r}_m^{\ell o}$, $\mc{C}$, $\mc{M}$

\EndFunction
\end{algorithmic}
\end{algorithm}
\vspace{-10pt}

\subsection{Anchor Localization} \label{subsec:teach_anchor_initial}

Assume at time-step $k$ Anchor $i \notin \mc{C}_k$ is detected as the $\ell^\text{th}$ component of the sequence of detected anchors. In order to localize the anchor, an optimization problem is formulated using the predicted state $\mbfcheck{x}_k^\mathfrak{t}$, its uncertainty $\mbfcheck{P}_k^\mathfrak{t}$, and the interoceptive and exteroceptive measurements recorded over a short time window from $k$ to $k+\lambda$. The goal is to solve for the anchor position $\mbfhat{r}_m^{\ell o}$ to be added to the anchor map and the state estimates $\mbfhat{x}_{k:k+\lambda}^\mathfrak{t}$ to add $\mbshat{\pi}_{k:k+\lambda}^\mathfrak{t}$ to the estimated teach pose trajectory.

To formulate a discrete-time optimization problem over the states, the process models \eqref{eq:process_model_1}-\eqref{eq:process_model_clocks} are concatenated and discretized to yield a system process model of the form
\begin{equation}
    \label{eq:process_model_general}
    \mbf{x}_{k+1}^\mathfrak{t} = \mbf{f}_k \left( \mbf{x}_k^\mathfrak{t}, \mbf{u}_k^\mathfrak{t}, \mbf{w}_k^\mathfrak{t} \right),
\end{equation}
where $\mbf{u}_k^\mathfrak{t}$ represents the interoceptive measurements. Similarly, the height measurements \eqref{eq:height_meas} and range measurements \eqref{eq:range_meas_1s}-\eqref{eq:range_meas_js} at time-step $k$ are concatenated and represented using the discrete-time measurement model
\begin{equation}
    \label{eq:measurement_model_general}
    \mbf{y}_k^\mathfrak{t} = \mbf{g}_k \left( \mbf{x}_k^\mathfrak{t}, \mbf{r}_m^{a_i o}, \mbs{\nu}_k^\mathfrak{t} \right), \qquad i \in \mc{C}_k.
\end{equation}

The optimization problem is then formulated as a standard discrete-time batch problem with Gaussian errors, thus the \emph{maximum a posteriori} (MAP) estimate is the solution to
\begin{align*}
    &\left( \mbfhat{x}_{k:k+\lambda}^\mathfrak{t}, \mbfhat{r}_m^{\ell o} \right) \\
    &= \underset{\left( \mbf{x}_{k:k+\lambda}^\mathfrak{t}, \mbf{r}_m^{\ell o} \right)}{\operatorname{arg \hspace{2pt} min}} \norm{\mbf{x}_k^\mathfrak{t} - \mbfcheck{x}_k^\mathfrak{t}}^2_{\mbfcheck{P}_k^\mathfrak{t}} + \sum_{j=k}^{k+\lambda} \norm{\mbf{y}_{j}^\mathfrak{t} - \mbf{g}_j \left( \mbf{x}_j^\mathfrak{t}, \mbf{r}_m^{\ell o}, \mbf{0} \right)}^2_{\mbf{R}_j^\mathfrak{t}} \\ & +  \sum_{j=k}^{k+\lambda-1} \norm{\mbf{x}_{j+1}^\mathfrak{t} - \mbf{f}_j \left( \mbf{x}_j^\mathfrak{t}, \mbf{u}_j^\mathfrak{t}, \mbf{0} \right)}^2_{\mbf{Q}_j^\mathfrak{t}} + R_\text{h}^{-1} \left( h - \left[ 0 \hspace{4pt} 0 \hspace{4pt} 1 \right] \mbf{r}_m^{\ell o} \right)^2. 
\end{align*}

The last error term corresponds to a prior on the vertical distance of the anchor to the floor being $h$. This prevents the estimate from converging to the local minimum associated with the \emph{flip ambiguities} \cite{Moore2004} that would result in the anchor being initialized below ground level. The covariance ${R}_{\text{h}}$ can be tuned to tailor to the user's confidence in the height $h$.

The initial iterate of the states $\big(\mbfhat{x}_{k:k+\lambda}^\mathfrak{t}\big)^{(0)}$ is obtained by dead-reckoning, while the initial iterate of the anchor position $\big(\mbfhat{r}_m^{\ell m}\big)^{(0)}$ is obtained by assuming a height of $h$ and analytically solving for the other two components using the range measurements at time-step $k$. The iterates are updated using Gauss-Newton or Levenberg-Marquardt until convergence. Once the optimizer converges, the states $\mbshat{\pi}_{k:k+\lambda}^\mathfrak{t}$ are added to the estimated pose trajectory, and the current state of the state estimator is updated to be $\mbfhat{x}_{k+\lambda}^\mathfrak{t}$. Meanwhile, Anchor $i$ is added to the current set of active anchors $\mc{C}_{k + \lambda}$, and the ordered triple $(\mbfhat{r}_m^{\ell o}, i, \ell)$ is added to the anchor map $\mc{M}$.

\subsection{Trajectory Estimation} \label{subsec:teach_estimation}

The state estimator is a basic extended Kalman filter (EKF) that provides an estimate $\mbf{x}_k^\mathfrak{t} \sim \mc{N} \left( \mbfhat{x}_k^\mathfrak{t}, \mbfhat{P}_k^\mathfrak{t} \right)$ at every time-step $k \in \{0, \ldots, K\}$ to be stored as the trajectory to be tracked in the repeat pass. The process models \eqref{eq:process_model_1}-\eqref{eq:process_model_clocks} are linearized using a first-order Taylor series expansion, and discretized using a first-order hold \cite[Section 4.7]{farrell2008}.

If a new anchor is detected at time-step $k$, the state estimator does not correct the estimates at time-step $k$, but the predicted state $\mbf{x}_k^\mathfrak{t} \sim \mc{N} \left( \mbfcheck{x}_k^\mathfrak{t}, \mbfcheck{P}_k^\mathfrak{t} \right)$ is input into the anchor initializer. The state estimator then resumes predicting and correcting at time-step $k+\lambda$ with state estimate $\mbf{x}_{k+\lambda}^\mathfrak{t} \sim \mc{N} \left( \mbfhat{x}_{k+\lambda}^\mathfrak{t}, \mbfhat{P}_{k+\lambda}^\mathfrak{t} \right)$, which comes from solving the optimization problem formulated in Section \ref{subsec:teach_anchor_initial}. Additionally, when an encountered anchor is initialized and is active, the estimated anchor position is used in the correction step as if it is the true anchor position. 

\section{Repeat Pass} \label{sec:repeat_pass}

\subsection{State Initialization} \label{subsec:repeat_init}

At the end of the teach pass, the robot is assumed to land on the floor close to its original take-off location, such that the initial pose of the repeat trajectory is similar to that of the teach trajectory. However, it is unlikely that the robot lands at the same exact position and that the robot is oriented the same way it was when it first took off. Under a flat floor assumption, the robot would have the same height, pitch and roll, but the 2D position and heading are not necessarily the same. Therefore, before attempting to retrace the teach pass trajectory, the robot must estimate its initial 2D position, denoted $\mbftilde{r}_m^{z o, \mathfrak{i}} = \mbf{D} \mbf{r}_m^{z_0 o, \mathfrak{r}} \in \mathbb{R}^2$, where $\mbf{D} = \left[ \begin{array}{ccc}
        1 & 0 & 0 \\
        0 & 1 & 0
    \end{array} \right]$,
as well as its heading, denoted $\mbftilde{C}_{mb}^\mathfrak{i} \in SO(2)$. During the repeat pass's state initialization, the robot remains static, and therefore the process model associated with the 2D position and heading is simply $ \dot{\mbftilde{r}}_m^{z o, \mathfrak{i}} = \mbf{0}$, $\dot{\mbftilde{C}}_{mb}^\mathfrak{i} = \mbf{0}$.

As before, the robot still needs to estimate its biases and clock states. The process model of the biases and clock states are as given in \eqref{eq:process_model_biases} and \eqref{eq:process_model_clocks}, respectively. Therefore, the state vector to be initialized is of the form
\setlength{\tabcolsep}{55pt}
\begin{equation*}
    \mbftilde{x}^\mathfrak{i} = \big[ 
        (\mbstilde{\pi}^\mathfrak{i})^\trans \hspace{6pt} (\mbs{\beta}^{\text{acc},\mathfrak{i}})^\trans \hspace{6pt} (\mbs{\beta}^{\text{gyr},\mathfrak{i}})^\trans \hspace{6pt} (\mbf{x}^{\text{UWB},\mathfrak{i}})^\trans
    \big]^\trans,
\end{equation*}
where $\mbstilde{\pi}^\mathfrak{i} = \big[ \begin{array}{cc}
        \big(\mbftilde{r}_m^{z o,\mathfrak{i}}\big)^\trans & \tilde{\theta}_{mb}^\mathfrak{i}
    \end{array} \big]^\trans$.
    
The anchor-independent range measurements are of the same form as given in \eqref{eq:range_meas_j1}. However, the anchor-dependent measurements \eqref{eq:range_meas_1s} and \eqref{eq:range_meas_js} are a function of the robot's pose, and therefore are rewritten explicitly with the 2D states by isolating the components of the third dimension, thus yielding
\begin{align*}
    {y}_k^{p_1a_i, \mathfrak{i}} = \f{1}{c} &\sqrt{ n^1_k + m^1_k } + \f{1}{2} \tilde{\nu}_{k,1}^{1,R} + \f{1}{2} \tilde{\nu}_{k,2}^{1,R}, \\
    {y}_k^{p_ja_i, \mathfrak{i}} = \f{1}{c} &\sqrt{ n^j_k + m^j_k } + {\tau}_k^{p_j p_1, \mathfrak{i}} + \tilde{\nu}_{k,2}^{j,R} - \f{1}{2}\tilde{\nu}_{k,1}^{1,R} + \f{1}{2}\tilde{\nu}_{k,2}^{1,R}, \\
    \text{where} \qquad n^\ell_k &= \norm{\mbftilde{r}_m^{z_ko, \mathfrak{i}} - \mbf{D}\mbf{r}_m^{a_io} + \mbftilde{C}_{mb_k}^\mathfrak{i} \mbf{D} \mbf{r}_{b_k}^{p_\ell z}}^2, \\
    m^\ell_k &= \norm{ \mbf{E}\mbf{r}_{m}^{p_\ell z} - \mbf{E}\mbf{r}_m^{a_i o}}^2,
\end{align*}
$j \in \{2,3\}$, $\ell \in \{1,2,3\}$, $E = [
    0 \hspace{3pt} 0 \hspace{3pt} 1 ]$, and $\mbf{E}\mbf{r}_{m}^{p_1z} = \mbf{E}\mbf{r}_{b_k}^{p_1z}$ since $\mc{F}_m$ is fixed to the initial body frame of the teach pass.

The height measurements are not useful when the robot is static on the floor, but the accelerometer and gyroscope measurements can be used as bias measurements as any non-zero measurement is a direct consequence of only gravity, noise, and biases. In fact, in the static case, the accelerometer and gyroscope measurements are modelled as 
\begin{equation}
    \mbf{y}_b^{\text{acc},\mathfrak{i}} = -\mbs{\beta}_b^{\text{acc},\mathfrak{i}} - \mbf{g}_b - \mbf{w}_b^{\text{acc},\mathfrak{i}}, \quad \mbf{y}_b^{\text{gyr},\mathfrak{i}} = -\mbs{\beta}_b^{\text{gyr},\mathfrak{i}} - \mbf{w}_b^{\text{gyr},\mathfrak{i}}
\end{equation}
where again assuming that the robot lies on a flat ground and that $\mbf{g}_w = [ \begin{array}{ccc}
    0 & 0 & -g
\end{array} ]$, thus $\mbf{g}_b = \mbf{g}_w$.

In order to obtain a reliable estimate of the initial pose of the robot, a batch estimation problem similar to the one presented in Section \ref{subsec:teach_anchor_initial} is formulated. As before, a concatenated and discretized process model and measurement model of the range, accelerometer, and gyroscope measurements of the form 
\begin{equation}
    \mbftilde{x}_{k+1}^\mathfrak{i} = \mbftilde{f}_k \left( \mbftilde{x}_k^\mathfrak{i}, \mbftilde{w}_k^\mathfrak{i} \right), \qquad  \mbftilde{y}_k^\mathfrak{i} = \mbftilde{g}_k \left( \mbftilde{x}_k^\mathfrak{i}, \mbf{r}_m^{a_1 o}, \mbstilde{\nu}_k^\mathfrak{i} \right) 
\end{equation}
are assumed, where $k \in \{ 0, \ldots, L\}$.

\noindent The initialization problem is then formulated as
\begin{align*}
    \hat{\mbftilde{x}}_{0:L}^\mathfrak{i} = \arg \min_{\mbftilde{x}_{0:L}^\mathfrak{i}} & \norm{\mbftilde{x}_0^\mathfrak{i} - \check{\mbftilde{x}}_0^\mathfrak{i}}^2_{\check{\mbftilde{P}}_0^\mathfrak{i}} +  \sum_{j=0}^{L-1} \norm{\mbftilde{x}_{j+1}^\mathfrak{i} - \mbftilde{f}_j \left( \mbftilde{x}_j^\mathfrak{i}, \mbf{0} \right)}^2_{\mbftilde{Q}_j^\mathfrak{i}} \\ &\hspace{45pt} +  \sum_{j=0}^{L} \norm{\mbftilde{y}_{j}^\mathfrak{i} - \mbftilde{g}_j \left( \mbftilde{x}_j^\mathfrak{i}, \mbfhat{r}_m^{1o}, \mbf{0} \right)}^2_{\mbftilde{R}_j^\mathfrak{i}}.
\end{align*}
This relies on an assumption that exactly one anchor is within communication range at the beginning of the trajectory, and the position of the robot is initialized relative to that anchor's estimated position $\mbfhat{r}_m^{1 o}$.

As before, Gauss-Newton or Levenberg-Marquardt can be used to solve this problem. The estimate $\hat{\mbftilde{x}}_L$ is then used to initialize the repeat pass's EKF. In particular, 
\begin{equation*}
        \mbfcheck{x}_0^\mathfrak{r} = \big[ (\check{\mbs{\pi}}_0^\mathfrak{r})^\trans \hspace{5pt} (\hat{\mbs{\beta}}_L^{\text{acc}, \mathfrak{i}})^\trans \hspace{5pt} (\hat{\mbs{\beta}}_L^{\text{gyr}, \mathfrak{i}})^\trans \hspace{5pt} (\mbfhat{x}^{\text{UWB},\mathfrak{i}}_L)^\trans \big],
\end{equation*}
where $\check{\mbs{\pi}}_0^\mathfrak{r} = \big[ 
        \big(\hat{\mbftilde{r}}_m^{z_L o,\mathfrak{i}}\big)^\trans \hspace{9pt} 0_\text{h} \hspace{9pt} \mbf{0}_\text{v}^\trans \hspace{9pt} \mbf{0}_\text{rp}^\trans \hspace{9pt} \hat{\tilde{\theta}}_{mb_L}^\mathfrak{i}
    \big]$.
The components $0_\text{h} \in \mathbb{R}$, $\mbf{0}_\text{v} \in \mathbb{R}^3$, and $\mbf{0}_\text{rp} \in \mathbb{R}^2$ correspond to an estimate of zero height, velocity, and roll and pitch, respectively.

\subsection{State Estimation}



    
    


    
    
    
    




\begin{figure*}
	\centering
	\begin{minipage}{0.6\textwidth}
		\centering
        \includegraphics[trim=1.6cm 3.7cm 1.2cm 2.2cm, clip, width=\textwidth]{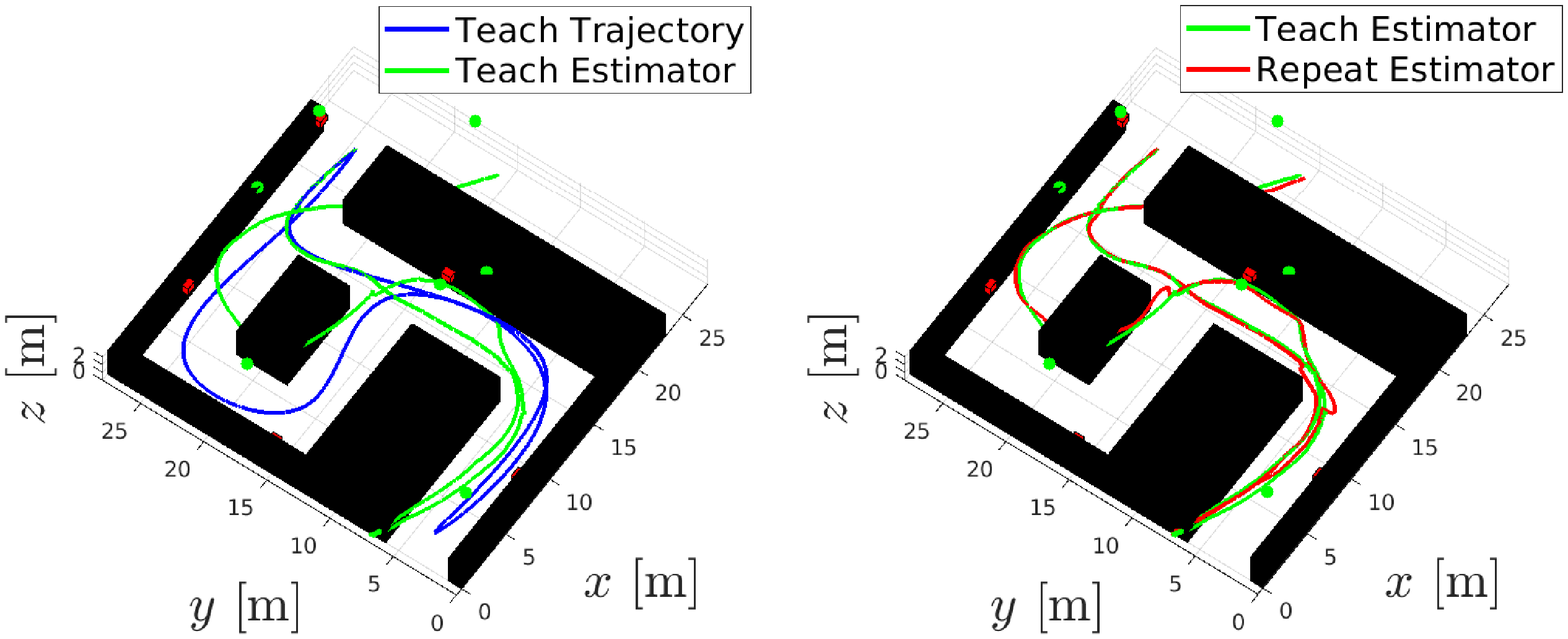}
        \caption{(Left) The true and estimated trajectories during the teach pass, and the true and estimated anchor positions as represented by the red and green blocks, respectively. There are more green blocks than red blocks as an anchor is initialized every time it is encountered. (Right) The estimated teach and repeat trajectories. }
        \label{fig:single_run_TandR}
	\end{minipage}\hspace{5pt}
	\begin{minipage}{0.34\textwidth}
    	\centering
        \includegraphics[width=\columnwidth]{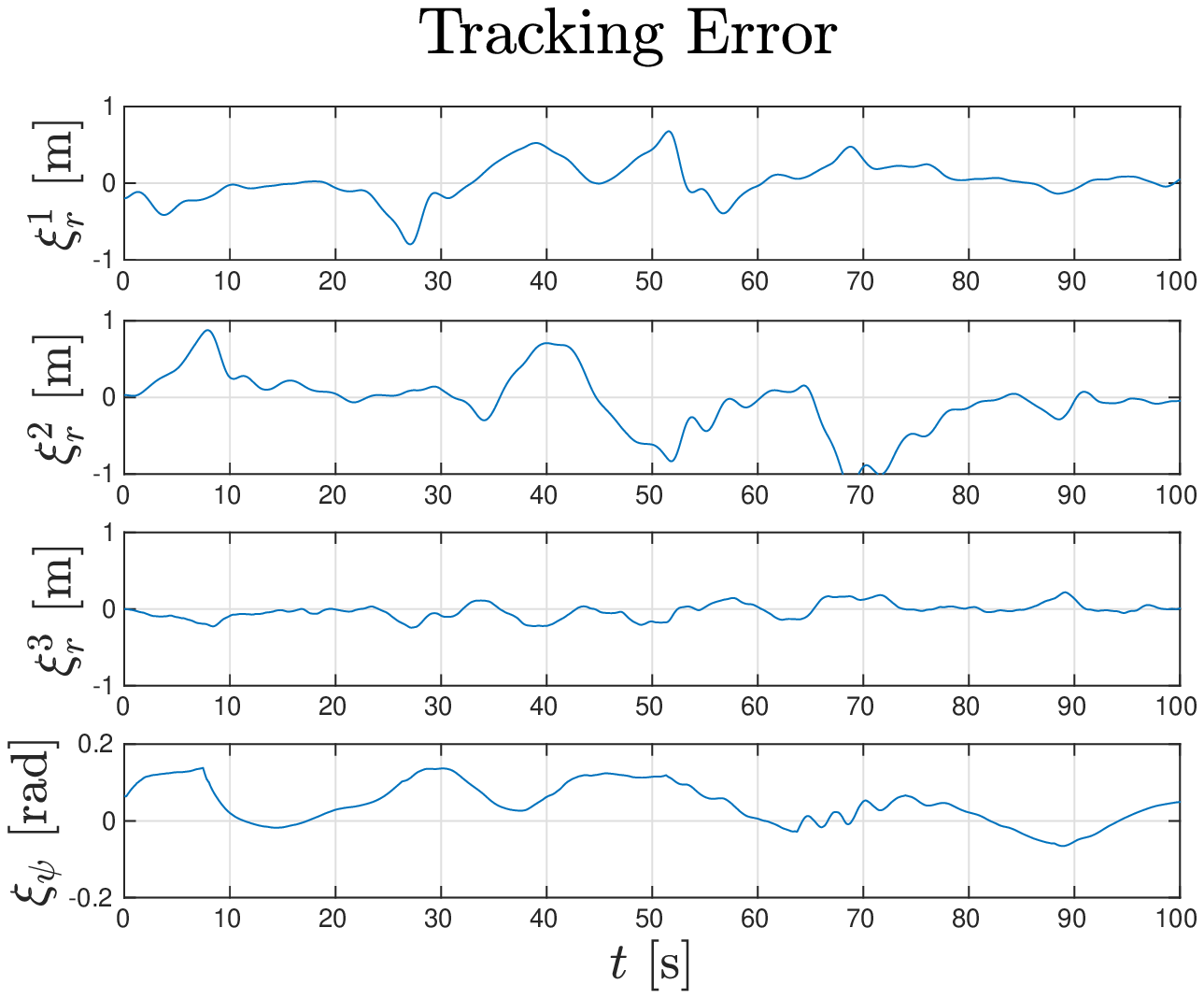}
        \caption{The position and heading error of the true repeat pass trajectory relative to the true teach pass trajectory.}
        \label{fig:tracking_err}
	\end{minipage}\hspace{8pt}
	\vspace{-7pt}
\end{figure*}

The state estimator in the repeat step is an EKF with the same model assumptions as the EKF used in the teach step. The repeat step EKF provides an estimate $\mbf{x}_k^\mathfrak{r} \sim \mc{N} \left( \mbfhat{x}_k^\mathfrak{r}, \mbfhat{P}_k^\mathfrak{r} \right)$ at every time-step $k \in \{0, \ldots, K\}$. The main differences between the teach and repeat estimators are that the state estimate and its uncertainty are initialized using the algorithm discussed in Section \ref{subsec:repeat_init} for the repeat estimator, and that the anchor position estimates from the teach step are assumed to be the true position of the anchors.   

A key component of the repeat pass is the anchor sequence tracker that allows the robot to match each anchor with the corresponding feature from the anchor map. This requires the robot to keep track of active anchors in a manner similar to the teach pass estimator, as presented in Section \ref{sec:teach_pass}. The tracker removes anchors that are no longer within communication range from the set of active anchors, and increments through the anchor map as new inactive anchors are detected.

	

			
			
	

\subsection{LQR Controller}

The trajectory-tracking controller implemented is based on the finite-horizon LQR controller presented in \cite{Cohen2020}. The trajectory to be tracked is the teach pass estimated pose trajectory $\mbshat{\pi}^\mathfrak{t}$, represented at any time-step $k$ as an element of a matrix Lie group $\mbfhat{X}_k^\mathfrak{t} \in SE_2(3)$. Similarly, the repeat pass estimated pose $\mbshat{\pi}_k^\mathfrak{r}$ at time-step $k$ is represented as $\mbfhat{X}_k^\mathfrak{r} \in SE_2(3)$. The left-invariant tracking error is then $\mbfdel{X}_k = \big(\mbfhat{X}_k^\mathfrak{t}\big)^{-1} \mbfhat{X}_k^\mathfrak{r}$. 

The control inputs at any time-step $k$ are the thrust $f_{b_k}$ and the angular velocity $\mbs{\omega}_{b_k}^{b_k w}$. The dynamics are modelled as
\begin{align}
    \mbfdot{v}_m^{zo / w} &= \mbf{C}_{mb} \left[ \begin{array}{ccc}
        0 & 0 & f_k
    \end{array} \right]^\trans + \mbf{g}_m + \mbf{w}_m^f, \\
    \mbfdot{C}_{mb} &= \mbf{C}_{mb} \left( \mbs{w}_{b}^{bw} + \mbf{w}_b^\omega \right)^\times,
\end{align}
and the command inputs are computed similarly to \cite{Cohen2020}. 

Even though the estimated teach trajectory drifts from the true trajectory, the anchors are localized based on the drifted trajectory. Since the estimated anchor positions are utilized in the repeat pass estimator, the estimated repeat pass trajectory drifts in the same direction. By controlling the robot to retrace the estimated teach pass trajectory through knowledge of only the repeat pass estimator's state, the robot is therefore tracing the true teach trajectory. However, when there are no anchors within communication range, the repeat pass estimator's state might drift in a different direction than the teach pass estimator. During such intervals, the repeat pass estimator must not be trusted by the controller. Therefore, whenever no anchors are within range, the controller is temporarily turned off and the reference inputs from the teach pass are implemented. 

\section{Results} \label{sec:results}




The proposed UWB T\&R algorithm is evaluated in simulation in the environment presented in Figure \ref{fig:env} and with a 3-tag quadcopter. All sensor measurements are corrupted with Gaussian white noise with characteristics corresponding to standard low-cost sensors. The results of the teach and repeat passes are shown in Figures \ref{fig:env} and \ref{fig:single_run_TandR}. The teach pass estimator drifts with time, consequently also resulting in poor anchor localization. Nonetheless, the map is ``locally consistent" in the T\&R sense, meaning that locally the estimated trajectory relative to the estimated anchor position is representative of the true trajectory relative to the true anchor positions. Therefore, by tracking the estimated teach trajectory and using the estimated anchors in the repeat estimator, the LQR controller is able to to track the true teach trajectory as shown in Figure \ref{fig:single_run_TandR}. This is despite the robot having a poor and globally inconsistent estimate of the true trajectory.

The trajectory-tracking controller computes its tracking error at any time-step by comparing the teach estimator's pose and the repeat estimator's pose at that time-step. Therefore, the controller attempts to get the estimated repeat trajectory as close as possible to the estimated teach trajectory. Nonetheless, the controller does not know what its true tracking error is, which is the error between the true repeat trajectory and the true teach trajectory. This error shown in Figure \ref{fig:tracking_err} for the position and heading is the metric in which the proposed algorithm is evaluated. As shown for this particular example, the robot manages to autonomously retrace the teach trajectory to within a metre and approximately 10 degrees in heading.


\begin{figure}
    \centering
    \includegraphics[trim=0cm 0cm 1cm 0.2cm, clip,width=\columnwidth]{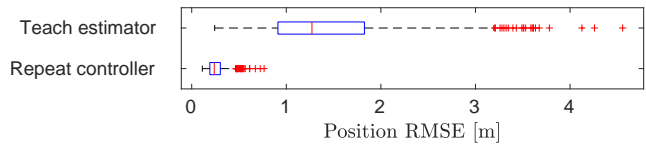}
    \caption{A box plot for the RMSE on 1000 Monte Carlo simulation trials. Although the teach estimator performs poorly, the robot manages to retrace the teach trajectory to within 1-metre accuracy for all runs.}
    \label{fig:boxplot}
\end{figure}

To statistically evaluate the performance of the proposed framework in simulation, 1000 Monte Carlo trials are performed. All trials share the same environment and the same teach trajectory, but the noise realizations, biases, clock states, and initial repeat pass's pose are distinct. The repeat position tracking root-mean-squared-error (RMSE) given by $\mbf{e}^{\text{RMSE},\mathfrak{r}} = \sqrt{\f{1}{N}\sum_{k=1}^K \norm{\mbf{r}^{z_k o,\mathfrak{r}}_m - \mbf{r}^{z_k o,\mathfrak{t}}_m}^2}$,
and the position estimation RMSE given by $\mbf{e}^{\text{RMSE},\mathfrak{t}} = \sqrt{\f{1}{N}\sum_{k=1}^K \norm{\mbfhat{r}^{z_k o,\mathfrak{t}}_m - \mbf{r}^{z_k o,\mathfrak{t}}_m}^2}$ are computed for each trial. The results are summarized in Figure \ref{fig:boxplot}. Despite the estimated trajectory available to the robot being poor with an average RMSE of 1.44 m, the tracking RMSE is always below 1 m, with an average of 0.26 m.

\section{Conclusion} \label{sec:conclusion}

UWB-based T\&R, as opposed to vision-based T\&R, is presented. A custom ranging protocol allows a 3-tag robot to range with the UWB anchors without compromising ranging frequency, and the measurements are fused with IMU and height measurements to compute a locally-consistent map and an estimated teach trajectory. To retrace the trajectory autonomously, the robot first estimates its initial pose in the repeat pass by remaining static, then fuses all on-board sensors with the map information from the teach pass as it navigates the same trajectory. The proposed algorithm is shown in simulation to be capable of retracing the trajectory with sub-metre tracking error. Future work involves addressing the influence of pose-dependent UWB bias on the performance of the proposed T\&R algorithm, and validating the algorithm in real-world experiments on ground and/or aerial vehicles.  




\bibliographystyle{IEEEtran}
\bibliography{IEEEabrv,TAR_ICRA}

\end{document}